%% file: direct.tex
\documentclass[10pt,twocolumn,letterpaper]{article}

\usepackage{iccv}
\usepackage{times}
\usepackage{epsfig}
\usepackage{graphicx}
\usepackage{amsmath}
\usepackage{shortbold}
\usepackage{amssymb}
\usepackage{caption}
\usepackage{booktabs}
\usepackage{epigraph}
\usepackage{verbatim}
\usepackage{multirow}
\usepackage[usenames,dvipsnames]{color}

\usepackage[pagebackref=true,breaklinks=true,letterpaper=true,colorlinks,bookmarks=false]{hyperref}

\iccvfinalcopy 

\def\iccvPaperID{603} 

\ificcvfinal\fi

\begin{document}
\newcommand{\note}[1]{\textcolor{red}{#1}}

\title{In Defense of the Direct Perception of Affordances}

\author{David F. Fouhey,~~~~Xiaolong Wang,~~~~Abhinav Gupta \\
The Robotics Institute, Carnegie Mellon University}

\maketitle

\begin{abstract}
The field of functional recognition or affordance estimation from images has
seen a revival in recent years. As originally proposed by Gibson, the
affordances of a scene were directly perceived from the ambient light: in other
words, functional properties like sittable were estimated directly from incoming pixels. 
Recent work, however, has taken a mediated approach in which affordances are derived
by first estimating semantics or geometry and then reasoning about the affordances.
In a tribute to Gibson, this paper explores his theory of affordances as originally proposed. 
We propose two approaches for direct perception of affordances and show that they obtain
good results and can out-perform mediated approaches. We hope this paper can
rekindle discussion around direct perception and its implications in the
long term.  
\end{abstract}

\input{introduction.tex}

\begin{figure*}[t]
\includegraphics[width=\linewidth]{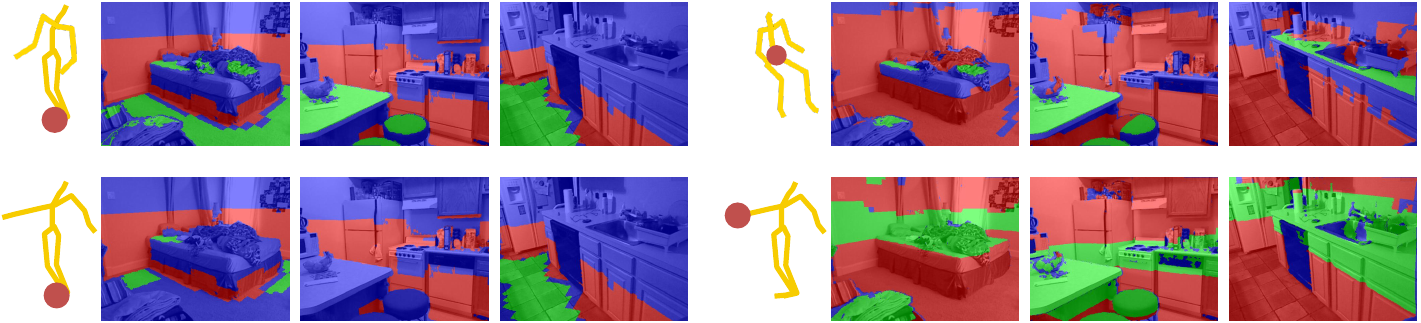} 
\caption{\small Sample labelings from our automatic method. The action and point of contact is indicated with a red dot (e.g., upper right
is the contact location of the pelvis when sitting upright).
\textcolor{ForestGreen}{Green: Supports Affordance}; \textcolor{red}{Red: does not}; \textcolor{blue}{Blue: unknown or missing data.}
Note that our method captures the {\it physical} affordances of the scene, so it permits standing on the counter and
on the stools in the middle example, even though we would not expect a human to do so. }
\label{fig:affLabels}
\vspace{-0.2in}
\end{figure*}

\section{Related Work}
\label{sec:related}
\input{relatedwork.tex}

\begin{figure*}[t]
\centering
\includegraphics[width=\linewidth]{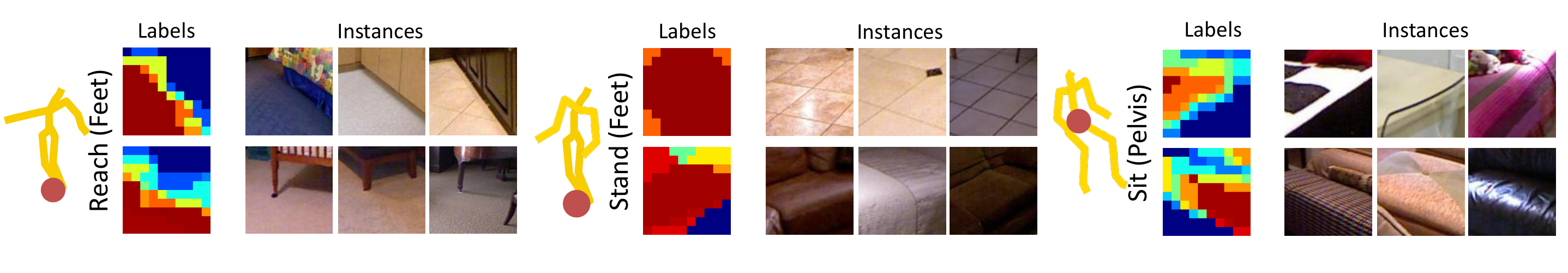}
\vspace{-0.4in}
\caption{Example affordance primitives learned by our method on the NYU Depth v2 dataset}
\label{fig:patchExamples}
\vspace{-0.2in}
\end{figure*}

\section{Overview}
\label{sec:overview}
\input{overview.tex}

\section{Generating Affordance Labels}
\label{sec:data}
\input{data.tex}

\section{Method}
\label{sec:learning}
\input{method.tex}

\section{Experiments}
\label{sec:experiments}

\begin{figure*}
\centering
\includegraphics[width=1.0\textwidth]{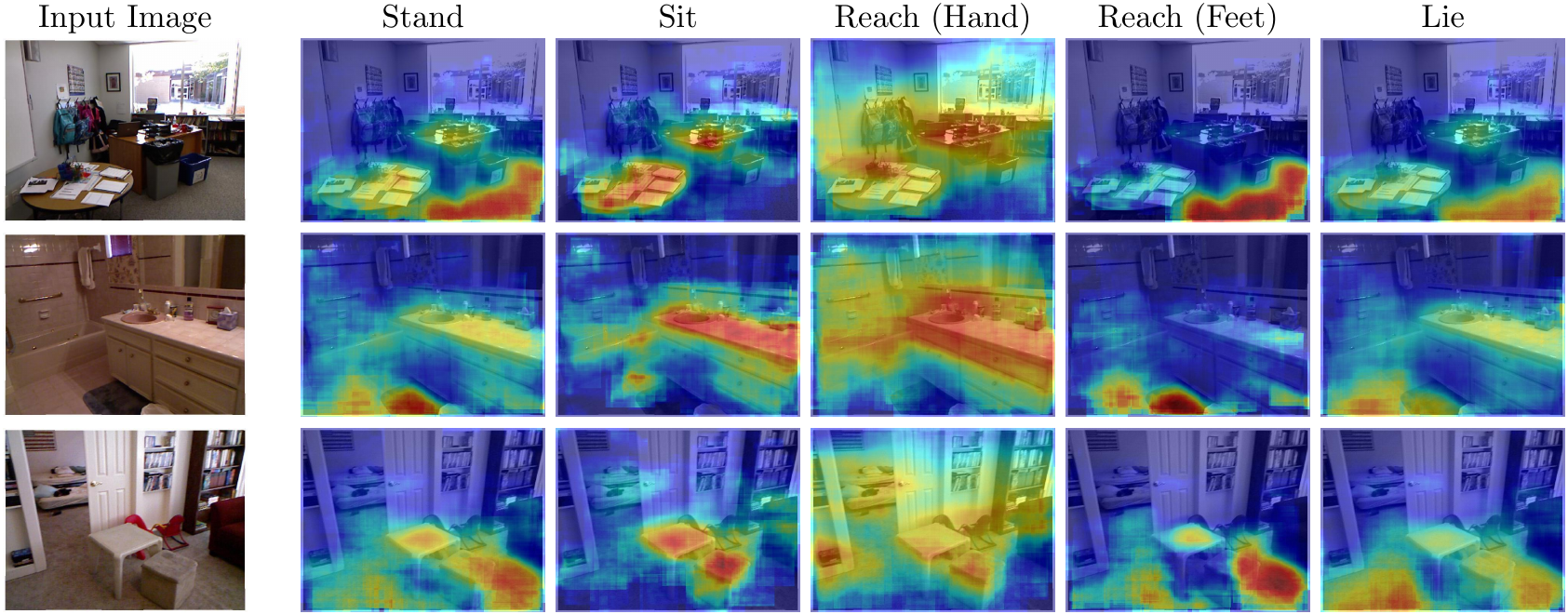} 
\caption{Sample results from our mid-level patch model 
(\textcolor{red}{red: supports affordance}; \textcolor{blue}{blue:does not}).
Our method captures distinctions between the affordances: for instance, one probably cannot 
touch anything standing on top of the desk in row 1.
}
\label{fig:respatch}
\vspace{-0.1in}
\end{figure*}

\input{experiments.tex}

\input{discussion.tex}

{\footnotesize \noindent {\bf Acknowledgments:} 
This work was partially supported by NSF IIS-1320083, ONR MURI N000141010934, 
Bosch Young Faculty Fellowship to AG and NDSEG fellowship to DF. 
The authors thank NVIDIA for GPU donations. 
The authors would like to thank Abhinav Shrivastava and Ishan Misra for helpful discussions.
}

{\small
\bibliographystyle{ieee}
\bibliography{shortstrings,local}
}

\end{document}

%% file: introduction.tex
\section{Motivation}

{\small
\epigraph{\em
The meaning or value of a thing consists of what it affords.}
{{\sc James J. Gibson},\\ The Ecological Approach to Visual Perception}}
\vspace{-.1in}

Functional recognition has had an amazing and rich history in psychology, robotics, and computer vision,
starting with early Gestaltist psychologists in the 1930s to recent data-driven computational techniques.
But one man deserves credit for functional recognition like no other: James J. Gibson. Gibson proposed the idea of
affordances\footnote{Gibson defined affordances as opportunities of
interaction in the scene} in the 1970s and tried to unify the perception of
space, objects and the meaning of objects. He single-handedly took 
charge and led a crusade against semantics, arguing that instead we should
use layout, 3D and functions as a way to perceive and understand the rich world
around us.

While the idea of affordances has always been appealing to 
psychologists and AI researchers, its computational implementation has always 
been murky. Early researchers tried using shape and other low-level features to
extract affordance properties from scenes, but these efforts failed due to an inability 
to reliably handle noisy data. However, in recent years, the field of functional recognition has
seen a revival in both psychology and the field of computer vision. The
resurrection of interest in computational affordances can be attributed to 
arrival of probabilistic approaches to reasoning under uncertainty. 
Recently, one common approach to estimating affordances in a given image is
to first obtain a semantic or 3D understanding of an image and then infer the affordances via reasoning.
This is referred to as a mediated approach, where semantics or 3D are the intermediate representation: for example, 
one can detect chairs in an image or predict a metric 3D reconstruction in order to label sittable surfaces.

\begin{figure*}
\includegraphics[width=\textwidth]{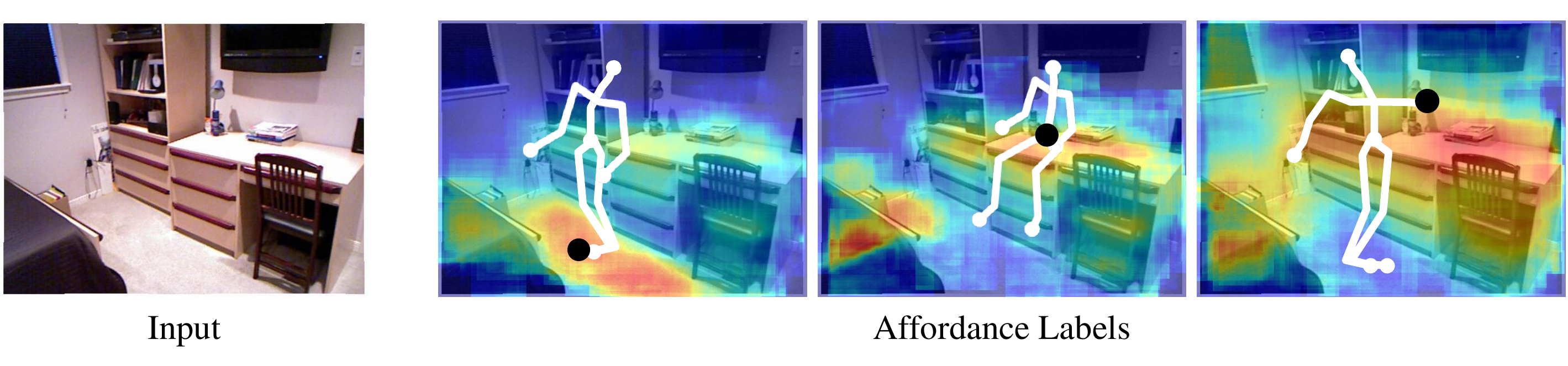}
\vspace{-0.3in}
\caption{When humans look at a scene, we can instantly recognize the scene's
{\it affordances}, or what we can do in it: we can stand on the floor and the
desk if necessary; we can sit on the desk too, but not on the bookshelf. 
We illustrate these by placing the appropriate human pose (white) and contact point
(black) on top of likely parts of a per-pixel functional map estimated by one of our algorithms.
This paper proposes to revisit the notion of recognizing functional properties
directly from pixel data. Note the humans are for illustrative purposes only; our approach generates the heatmap.
} 
\label{fig:teaser}
\end{figure*}

However, we have strong evidence to believe that Gibson himself would not be
completely satisfied with this development: most current approaches use reasoning to estimate
affordances. In other words, one first extracts a semantic or 3D understanding of the scene,
and then figures out the affordances from them. Gibson had a different view:
when he proposed affordances in the first place, he believed strongly in the idea of 
their direct perception. Gibson argued that affordance and surface layout are the
properties of the environment and that perception does not depend on
meaning-conferring inferences; instead, we simply gather information from
the ambient array of
light~\footnote{http://www.ffyh.unc.edu.ar/posgrado/cursos/chemero.pdf}.
Gibson's distaste for inferential perception is obvious when he defines
direct perception to be the central question in the theory of
affordances: {\em ``The central question for the theory of affordances is
not whether they exist and are real but whether information is
available in ambient light for perceiving them''}~\cite{Gibson79}. He
further argues: {\em ``The hypothesis of information in ambient
light to specify affordances is the culmination of ecological
optics. The notion of invariants that are related to one
extreme to the motives and needs of an observer
... provides a new approach to psychology''}. 

Is it really possible to directly perceive affordances?
Can we really learn a classifier to find sittable surfaces when we cannot even
build a chair detector yet? Do affordances provide another set of invariants
with respect to the environment? After 35 years of Gibson's theory of affordances,
we revisit the idea of doing direct perception and explore affordances
as they were originally proposed. 

In a tribute to Gibson, we show that it is indeed possible to have algorithms 
that directly perceive affordances in a scene and we evaluate why the direct approach
may have advantages over mediated approaches. To do this, we
present two approaches that estimate affordances without using an intermediate 
representations, one based on mid-level representations and the other based on
deep learning. We show that these approaches can out-perform mediated approaches based on
estimating an intermediate representation in the form of semantics or 3D.
Additionally, we present evidence that shows that our affordances can act as a 
feature for 3D scene understanding.

This paper is a proof of concept that aims to rekindle discussion of direct 
perception. Our ultimate goal is to obtain an action-centric understanding of scenes. 
It is genuinely not clear whether direct perception is the right approach 
for getting this understanding: there are
many reasons why direct perception may be wrong, and the final answer may
involve mediated perception. However, we believe that the direct approach deserves 
a genuine second look.

\subsection{A Modern View of Direct Perception}

Gibson was never clear about what he meant by direct
perception. Ullman, in his classic work {\it Against Direct Perception}
\cite{Ullman80} gives a good definition: the representation cannot be
decomposed further at that particular level of representation. He also provided
an illustrative example:  adding two single digit numbers is a
direct (via lookup), but multi-digit addition is not, because it is decomposed
into these single digit operations. The ``direct'' lookup for single digits
might be electrically implemented as a complex operation, but this is at a
lower level of understanding than addition.

Building on this idea, we try to define what we mean by
direct perception and how it differs from most recent work on affordances in
computer vision. By direct perception, we mean that we have a
{\it feedforward}  mapping from pixels to our desired output in terms of a
function. This function could be linear or non-linear, use the whole image or part
of an image; it could be simple or the composition of several simple functions,
\etc. However there are a few properties that must be satisfied: (a) In the
process, the function {\it cannot} calculate some externally meaningful 
representation (i.e., pre-specified and immediately human-interpretable)
as an intermediate representation, such as semantics or 3D
(b) The function cannot be defined over the output space itself but should be representable by the form $y = f(x)$.

For instance, if we were predicting semantic classes, a random forest on
bag-of-words features over dense SIFT on superpixels would be direct; CRF
reasoning on top would not be because it is not feedforward and does
reasoning over the labels. A template over HOG would be direct, but using
these detection in terms of categories to infer a scene class would not be
direct because it makes use of a semantically meaningful intermediate
representation.

In this paper, we demonstrate two direct perception approaches for affordances.
The first is a bank of linear templates over HOG features
\cite{Dalal05} that are used to directly transfer affordance labels. 
The second approach is inspired by Gibson's view of
affordances as an invariance. We learn a deep convolutional
neural network (CNN) over RGB images to create segmentation masks for affordances
like sittable, walkable, etc. It should be highlighted that while CNNs
have layers, they are feedforward. Furthermore, for every deep network
there also exists a ``shallow'' non-linear network which can often yield similar
performance ~\cite{Ba14}. This suggests that intermediate layers are just for
convenience of learning and representation.

%% file: relatedwork.tex
The notion that the objects are defined by their functions (e.g., ``sitting'')
dates back to the early 20th century when Gestaltists proposed that some
functions of objects can be perceived directly. These ideas were picked
upon by James J. Gibson who proposed the theory of affordances~\cite{Gibson79}.
Affordances can be seen as ``opportunities for interactions'' provided by the
environment, which can be perceived directly from form and shape. While
this notion of affordances inspired a lot of work in psychology and related
fields, it took the back burner because of multiple counter examples to the
theory of affordances (e.g., the famous ``mailbox vs. trash can'' in \cite{Palmer99}).

Inspired by Gibson's ideas and possible utility in robotics, the field of
computer vision has time and again fiddled with the idea of functional
recognition~\cite{Stark, Rivlin}. Most approaches were based on
recognition techniques that first estimated physical attributes/parts and then
used these to reason about affordances. For example. \cite{Stark, Binford} used
manually-defined rules to reason about 3D objects and predict affordances. 
Unfortunately, none of these approaches scaled up due to a lack of data, inability
to learn from data, and inability to handle noise/clutter in the input scenes. 

In the last decade, as we made substantial progress in the field of
semantic and 3D scene understanding, the topic of functional recognition has
started to gain relevance once again. The major gains in the semantic and 3D
scene understanding were due to the result of availability of large scale
training data and better classifiers that can be trained using this data.
However, this approach of training direct classifiers did not seem to result in
significant gains for the field of functional recognition. This is primarily
due to lack of large training datasets for affordances. While it is easy to
label objects and scenes, labeling affordances is still manually intensive.
Instead, the field has focused on using reasoning on top of semantic and 3D
scene understanding to obtain affordance results. For example, \cite{Gupta07}
and \cite{Kjellstrom08} focused on using relationships between objects and
actions to improve each other. Yao et al.~\cite{Yao10a} took this idea a
step-further and modeled relationships between objects and poses.
\cite{Delaitre12} proposed an approach to learn these object-pose
relationships via time-lapse videos themselves but still used semantic classes
of objects as intermediate representation. Recently, \cite{Zhu14} et al.
proposed a way to reason about object affordances by combining object
categories and attributes in a knowledge base manner.
One exception is the use of estimated grasps as a feature
for object classification in \cite{Castellini11}, which focuses
up-close objects and maps to a set of recorded hand grasps. Our work addresses scene-level
pose categories and generates labels from 3D point-clouds.

There has also been efforts to link 3D scene understanding with 
affordances~\cite{Gupta11, Grabner11, Fouhey12, Zhao13, Xie13}, often with
the goal of estimating affordances. For example,
Gupta et al.~\cite{Gupta11} use indoor scene understanding approaches,
followed by person fitting to obtain affordances. Similar examples were used
to define object categories such as chairs~\cite{Grabner11}. \cite{Zhao13}
designed an AND-OR graph to link 3D understanding and affordances to better
understand scenes. There have also been efforts to use RGBD data itself to
predict affordances at a scene level~\cite{Jiang13,Koppula14,Koppula15}
and object level~\cite{Zhu15}.

In contrast to most prior work, our goal is to explore the problem 
of estimating affordances from a different perspective: a direct, rather than mediated, approach. 
To do so, we overcome the data-barrier for learning affordances and present the
first mid-level or deep learning approach to estimate affordances without any
semantics or 3D reasoning.

%% file: overview.tex
Our goal is to investigate whether it is possible to learn to directly perceive
affordances from an input image. That is, can we learn a classifier to predict
the affordance regions such as: ``sittable'', ``walkable'', etc? 
One bottleneck in the past has been a lack of data, so in 
Section~\ref{sec:data}, we introduce a technique to generate labels for 
affordances: we analyze the 3D structure to determine where humans can 
physically perform four actions:
sitting, walking, reaching, and laying. 

Given this data, we investigate two approaches for learning to infer these
affordances on new data in Section~\ref{sec:learning}. One is based on 
mid-level patch representations; the other is a convolutional neural net. 
In Section \ref{sec:experiments}, we evaluate both of these methods 
on both our data generated on the challenging NYUv2 dataset \cite{Silberman12} 
and on hand-labeled affordances. Finally, we report proof-of-concept
results showing that direct affordances can aid 3D understanding.

%% file: data.tex
Learning a direct mapping from image pixels to affordances requires large
amounts of training data. Unfortunately, collecting such data using manual
labeling is expensive due to extensive labor required for the task.
At first glance, one possibility is to map existing semantics or 3D labels to
affordances. However, we cannot straightforwardly repurpose existing semantic or
3D labels: bounding boxes or segments might tell us that there is a couch in a
region, but they do not tell where we can sit on the couch; similarly, surface
normals might tell us that a surface faces upwards, but do not say whether
there is enough free-space in front or above for a person to sit.

To solve this, we note that past work \cite{Gupta11} have proposed a way to use human poses 
and 3D scene data to estimate affordance regions. Therefore, 
we apply the qualitative pose approach introduced in \cite{Gupta11} to the NYU Depth v2 dataset \cite{Silberman12}.
We begin with a depthmap, voxelize and complete it, and then check whether a 
human can physically take the given pose at each voxel with the appropriate 
support. For instance, to determine whether a location affords sitting, 
we check whether the pelvic joint has support from beneath and there is 
sufficient freespace in front and above.

The first step of our pipeline is obtaining a voxel representation from a depthmap which 
we will use for reasoning about the human pose compatibility with the scene at each location. 
The challenge is that sensors such as the Kinect only give a noisy 2.5D interpretation rather than an accurate 3D mesh: 
many parts of the scene are occluded or missing due to sensor noise
and the image only captures a limited field of view.
We obtain results with a simple approach that operates on a gravity-aligned Manhattan-world occupancy
voxel grid. Each cell in our grid is marked as either filled or not. We extract gravity using the method 
from \cite{Gupta13}, or from the floor when semantic labels are available; we then extract 
the remaining directions with a RANSAC-like approach. Once we have a coordinate frame, we 
backproject the depthmap to a grid with cube voxels of size $10 \textrm{cm}$. This gives
a one voxel-thick shell around the camera with the occluded regions unfilled. We fill the 
grid under the assumption that any non-downwards facing voxels (more than $45^\circ$) are supported 
from beneath. 

Now that we have a a voxel grid, we mark each voxel as supporting or not supporting the
affordance. We can take poses as in \cite{Gupta11} and convolve them
in the grid. As in \cite{Gupta11}, this is implemented by convolving four rotations of a 
3D filter that checks for freespace and occupancy criteria followed by support
criteria at the joint in question. This can be implemented efficiently as convolution by filters.
For instance, a standing filter would look for parts of the voxel grid that have: (1) a solid voxel 
beneath; (2) enough freespace above to fit an adult human above; (3) and are facing upwards. 
There are two large difficulties with this approach: large amounts of data relevant to our analysis
may be missing because it is outside the field of view; and even inside the field-of-view, the sensor 
data has holes and is noisy. Therefore, rather than hard-threshold the filter response, we mark voxels with
insufficiently strong of a filter response as unknown. These voxels are then backprojected to the image 
to give the final per-pixel labels. When available and applicable, we also use floor
labels to better filter the labels.

We apply our method to four poses: sitting upright (i.e., with legs in front as if on a chair), standing,
lying, and reaching out. For reaching, we compute affordances for both the hands (i.e., what can be touched)
and the feet (i.e., where one can put one's feet and touch). Our method obtains good results on most parts 
of most scenes. We show some example scenes in Fig.\ \ref{fig:affLabels}. We note a few failure modes: 
many stem from not always reasoning correctly about regions that are occluded or 
outside the field of view or from the discretization of our problem. Similarly, regions with large amounts of 
missing data pose challenges.

%% file: method.tex
Now that we have data, we can learn a models to estimate affordances.
In this section, we introduce two approaches for doing so.
Both are feed-forward, do not calculate any meaningful 
intermediate representations, and do not reason over the label space. 
Our experiments show that both achieve performance on a challenging dataset.

The first is based on local discriminative templates: we learn a bank of
filters that immediately imply a patch label at detection locations. Learning takes
the form of discriminative clustering and inference is simple label-transfer.
The second is based on convolutional neural nets (CNNs) and
maps an entire image to an image-level grid of affordances.

The approaches are deliberately simple to demonstrate the feasibility of 
directly predicting affordances: we treat the problem as $k$-way 
binary classification and do no post-processing of our output, for instance. 

\subsection{Approach 1: Mid-Level Elements}
We first build on existing work on discriminative mid-level patch mining
to learn a collection of templates over HOG features that directly predict which parts 
of a window in the image afford a particular action. For every affordance, we build a
patch-based model in the style of \cite{Fouhey13a}. 
These patches predict an affordance label wherever they have a high response. 

\noindent {\bf Patch Discovery:}
The goal is to discover a set of primitives that immediately tell us their affordance label in new
images. We learn a set of primitives for each affordance class by optimizing
two properties: (a) the primitive should be visually discriminative since
this allows for direct perception and requires no reasoning;
(b) the primitive should be informative about the underlying affordance.
Some elements satisfying this property are shown in  Fig.\ \ref{fig:patchExamples}.

Each element is defined in terms of: a detector $\wB$ over HOG features, which recognizes
the element in new images; a canonical form $\FB$, which represents the underlying
functionality in terms of how likely it is that each part of the patch permits
that action; and an instance indicator vector $\yB$, which indicates which patches
in the training set belong to the primitive. Similar to \cite{Fouhey13a}, our goal is 
to find $(\wB,\yB,\FB)$ that minimizes the following objective:
\begin{equation*}
\min_{\wB,\yB,\FB}{R(\wB) + \sum{L(\wB,\xB_i^A) + y_i\Delta(\FB,\xB_i^L)}}
\end{equation*}
subject to a cardinality constraint on the number of members $||\yB||_1$.
Here, $\xB_i^L$ represents a grid of affordance labels and $\xB_i^A$ represents
a patch's appearance features (HOG). The first and second term correspond to an 
SVM that maintains visual discriminativity. The final term penalizes inconsistent underlying functionality.

We solve for the patches with an iterative approach akin to \cite{Fouhey13a}:
we start by initializing each primitive's membership $\yB$ with a label-and-appearance clustering step 
over large numbers of randomly sampled patches. We then optimize each variable in turn:
if we have the patch's membership, for instance, we can figure out what part of the patch
supports sitting. We include the cross-validation step and background dataset from \cite{Doersch12,Singh12}.  
A few iterations gives a detector $\wB$ that we can run in new images as well as a label $\FB$ that can be
transferred. We obtain multiple primitives with multiple restarts.
A few patches are displayed in Fig.\ \ref{fig:patchExamples} along with their labels. Note each cluster has no single
semantic meaning and the 3D is often not the same at the pixel level.

We tailor the approach to affordance estimation. Two difficulties 
are that our labels exhibit severe class skew (e.g., only $3\%$ of the data is labeled as sittable) 
and have a relatively weak label space for clustering (binary masks).
We solve the first by initializing with patches sampled so that at least $25\%$ of the patch supports the
affordance; to solve the second problem, we cluster with multi-restart k-means on HOG, setting aside consistent clusters on each restart. 

\noindent {\bf Inference:}
Given a collection of detectors and their canonical forms, the inference procedure
is simple label transfer. We run all the detectors in a multi-scale fashion and simply transfer
the canonical form $\FB$ to the corresponding detection region, resizing as appropriate.
The final output is then an average of these canonical forms, weighted
by their Weibull calibrated \cite{Scheirer2012} detection score.

\subsection{Approach 2: CNNs}
\label{sec:CNN}
Gibson's view of affordances was based on invariances.
Interestingly, interest in invariances has had a resurgence in
recent years due to the strong performance of CNNs on many
vision tasks. These CNNs can be viewed as learning invariances
at different levels of representation. Therefore, we try a CNN-based 
model for predicting affordances. As in the mid-level approach, 
we treat the problem as a series of $k$ binary classification problems
and adapt an existing approach.

\noindent {\bf Representation and Learning:} 
Our approach takes an image and outputs a binary segmentation mask 
indicating which parts of the image, if any, support the affordance.
For simplicity, we adapt a standard architecture \cite{Krizhevsky12},
to this task: rather than predict a final soft-max output, we output
a $m \times n$ grid of binary logistic predictions. Our loss function is then
\begin{eqnarray}\label{eq:multilr}
\tiny
L(\xB,\yB) = -\left( \sum^{m \times n}_{i=1} y_i \log f_i(\xB) +  (1 - y_i) \log(1- f_{i}(\xB) ) \right). \nonumber
\end{eqnarray}
We set $m = n = 50$. We optimize the model via backpropagation
using standard stochastic gradient descent.

\noindent {\bf Inference:}
At test time, we pass the input image through the network and obtain a probability
at each pixel in our $m \times n$ grid; this is then upsampled to the 
input image size.

%% file: experiments.tex
We now report experimental evaluations done to investigate direct affordances.
Our experiments are guided by three questions. We want to see if (1) direct
perception is possible and how it compares to mediated
approaches; (2) how well our direct approach works on hand-labeled affordance data;
and (3) whether our affordance outputs are useful for 3D scene understanding.
We stress that the first point is non-obvious: many past works have insinuated
that estimating affordances directly is ineffective.

To investigate the first question, we try our approach on NYUv2 and compare
it with mediated methods based on using an intermediate semantic or 3D representation.
We then address the second question by trying a direct approach on hand-marked
affordances. Finally, we present two results that show
that our direct affordance estimates can be used as a feature to improve 
3D scene understanding.

\subsection{Setup}

\noindent {\bf Protocol:} We evaluate on the NYU v2 Depth dataset \cite{Silberman12}
using the standard train-test splits. For the CNN, which requires
large amounts of data to learn a representation, we use 100K frames
from the corresponding videos from the raw dataset for training.
Since we are interested in high-quality labels and our ground-truth generator
is an imperfect algorithm applied to noisy data, we evaluate quantitatively only on images for which
we have the best chance of labeling the data correctly. We ignore
the minority of scenes with no floor, which we use to get room extent. 

\noindent {\bf Evaluation Criteria:}
We treat affordance estimation as if it were a detection task, and compute
a precision-vs-recall tradeoff. We ignore pixels that our label 
generation approach is unsure of but obtained similar relative performance and
identical conclusions when these were included.
We summarize these curves with the area-under-the-curve.

\subsection{Mediated Perception Baselines}

\begin{table*}
\caption{Average precision on NYUv2 for affordance estimation. We report, left-to-right:
proposed direct approaches; mediated approaches using estimated intermediate representations; 
and mediated approaches using ground-truth intermediate representations. The best 
approach among non-ground-truth methods is consistently a direct one, often by a wide margin.}
\label{tab:results}
\vspace{-0.1in}
\centering
\begin{tabular}[b]{@{}l@{}p{2em}@{}cc@{}p{2em}@{}cc@{}p{2em}@{}c@{}p{4em}@{}ccc}\toprule
 & & \multicolumn{2}{c}{Direct} & & \multicolumn{2}{c}{via Semantics} & & \multicolumn{1}{c}{via 3D} & & \multicolumn{3}{c}{on Ground Truth} \\
 \cmidrule{3-4}\cmidrule{6-7}\cmidrule{9-9}\cmidrule{11-13}
Label & & CNN & Mid-level & & 13 Class & 40 Class & & Normals & & 13 Class & 40 Class & Normals \\
 \cmidrule{1-4}\cmidrule{6-7}\cmidrule{9-0}\cmidrule{11-13}
Standing &
 & \bf 88.55  & 83.81  & & 81.22  & 78.04  & & 82.07  & & 91.34  & 92.00  & \bf 93.08 
\\
Sitting Upright &
 & \bf 37.34  &  31.95  & & 28.07  & 25.08  & & 23.72  & & 47.98  & \bf 56.22  & 50.50 
\\
Reach Feet &
 & 68.40  &  \bf 71.70  & & 62.11  & 55.25  & & 65.83  & & 81.32  & 81.32  & \bf 82.48
\\
Reach Hand &
 & 50.70  & \bf 56.56  & & 55.16  & 53.11  & & 55.60  & & 66.80  & \bf 69.36  & 60.81 
\\
Lie Down &
 & \bf 82.10 & 76.04  & & 72.72  & 66.73  & & 74.38  & & 87.37  & 88.70  & \bf 90.37 
\\
\bottomrule
\end{tabular}
\end{table*}

The goal of this paper is to show the feasibility of estimating affordances directly, not to 
outperform mediated perception approaches.
However, to place our results in context and to examine the tradeoffs between 
mediated and direct approaches, we compare mediated approaches.
We build these approaches using two intermediate representations
common in the literature for affordance reasoning -- geometry and semantics. 

A mediated approach is defined by three things: the intermediate representation
(e.g., surface normals), the estimator for the 
representation (e.g., a normal estimator), and a decoder that 
maps the intermediate representation to the affordance. We now
define our intermediate representations, their estimators, and the decoder used by both.

\noindent {\bf Geometry:} Recent work \cite{Fouhey13a,Ladicky14b,Fouhey14c,Wang15}
has demonstrated the feasibility of estimating surface normals from a single image. We 
adopt these as our geometric representation. 
To learn a representation decoder, we discretize the
predictions into 40 classes using a codebook learned on the training data
following \cite{Ladicky14b,Wang15}. We use 3DP \cite{Fouhey13a} since it has training
code available.

\noindent {\bf Semantics:} We treat semantics as a pixel labeling problem. NYUv2 comes
with 849 detailed semantic labels which are typically condensed into a smaller set. 
We experimented with two sets: one \cite{Gupta13} has 40 classes the other \cite{Couprie13} has 13+unknown. 
To predict semantics, we use the automatic labeling environment (ALE), which 
implements an associative hierarchical CRF (AHCRF) discussed in \cite{Ladicky09,Ladicky10}. We 
use a common setup for all experiments.

\noindent {\bf Decoding Intermediate Representations:} We tried two ways
to turn intermediate representations into affordances and report results from the better performing one.
The simplest is a direct map (e.g., chair $\,\to\,$ sittable), which performed acceptably.
We were able to obtain better results by taking the surrounding context into consideration 
via a classifier over superpixels. We used a spatially-binned histogram
over the representation in a window around the superpixel as features for a random forest classifier.
All decoders are learned on cross-validated output.

\subsection{Results on NYU}

\begin{figure*}
\centering
\includegraphics[width=1.0\textwidth]{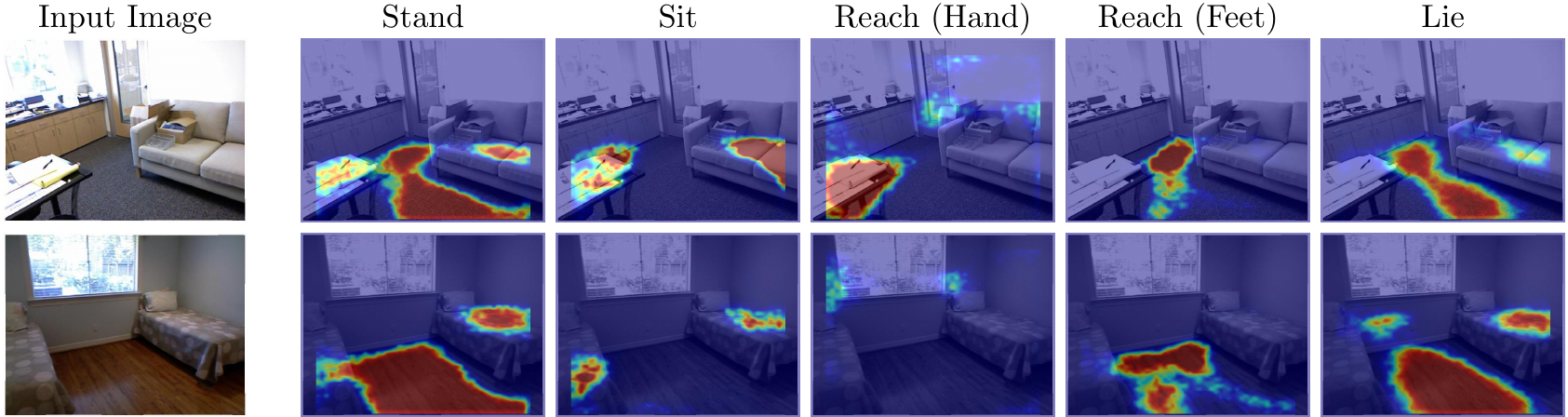}
\caption{Example results from our CNN
(\textcolor{red}{red: supports affordance}; \textcolor{blue}{blue: does not}).
We infer a good interpretation of affordances. Notice
areas where some affordances are possible but not others (e.g., stand vs. lie
in row 1, under the desk).}
\label{fig:resdeep}
\vspace{-0.1in}
\end{figure*}

We show some qualitative results of our mid-level and
CNN approaches in Figs.\ \ref{fig:respatch} and \ref{fig:resdeep} respectively. Since
our labels represent physical feasibility, we often predict actions that are {\it feasible}
but not culturally appropriate. For instance, we predict standing on tables and counters,
sitting on stoves and sinks, and so forth. 

Both models obtain good interpretations of most portions of the scenes. 
Our CNN based approach predicts more conservatively and with sharper boundaries than the Mid-Level approach. 
Both are often able to recognize affordances on objects that are themselves hard to parse, 
such as the sitting affordance in Fig.\ \ref{fig:respatch}:
our method detects these affordances even on the far-away desk in row 1 and small 
chair in row 2. While recovering the precise 3D might be challenging from a
single image, our method directly recognizes both as affording sitting simply
because they look as if they afford sitting. Notice that while the affordances
share some pixels in common, they are many differences: compare the parsings
of the desk in the top row of Fig.\ \ref{fig:respatch}: we can stand on it,
but we cannot reach to touch anything while standing there.
Similarly in the top row of Fig.\ \ref{fig:resdeep},
the CNN identifies that many areas where one can stand do not permit lying due
to a lack of free-space.

Typical failures include sometimes not getting the scale quite right
at places like edges; this may be where direct and intermediate approaches
based on 3D might have complementary strengths. Additionally, without
a global model, our local patch-based approach sometimes does not take
into enough context.

We report quantitative results in Table \ref{tab:results}.
The Mid-Level model is always better than the mediated approaches,
and the best approach is consistently a direct one, 
often by a large margin. The gap is smaller for some
classes for which there is an easy mapping from the intermediate representation
to the affordance (e.g., upwards-facing $\,\to\,$ standing), and larger for
those where the mapping is not as straight-forward (e.g., sitting). Curiously,
while the direct Mid-level method always out-performs the mediated approaches,
the best mediating representation varies from task to task: for sitting, it is semantics,
but lying it is 3D.

\par ~ \\
\noindent {\bf Why does mediated perception do poorly?} 
We now show evidence that suggests that the intermediate representation is itself to blame.
We first rule out the possibility that decoder or intermediate representation estimator 
(i.e., 3DP or AHCRF) is responsible for the difference between direct and indirect methods.
To absolve the decoder, we report results from identical models using
ground-truth intermediate representations: Table \ref{tab:results} shows that
these dramatically outperform using estimated representations. Thus,
large errors have already crept in before the decoder.
To absolve the estimator, we use the same machinery but skip the intermediate representation. 
For 3DP, this is equivalent to functional primitives; for semantics, this entails 
retraining the AHCRF, which consistently yielded substantial gains across all affordance categories and performance
roughly on par with the functional and CNN approaches.

The more likely culprit is then the intermediate representations themselves and how they cause and propagate 
errors. This can be checked by separately evaluating pixels depending on the whether their intermediate representation 
was correctly estimated. We find that correctly estimated pixels have a substantially higher performance than
incorrectly estimated ones. This can also be seen in performance across intermediate representations.
Intuitively, the richer 40-class representation should yield better results as it preserves more information,
and this is the case given ground-truth semantics. However, given estimated semantics, the 
coarse 13-class representation is always strikingly better as it makes fewer mistakes per-pixel.
This is only exacerbated by the intermediate classifier's ignorance of the ultimate task.
Confusing some categories (e.g., a night-stand and a table) is irrelevant for many functions but is as penalized as confusing a nightstand
and the wall. Thus, intermediate classification mistakes are often irreversible and do not permit
graceful degradation by the next stage.

\begin{figure}
\centering
\begin{tabular}{@{}c@{~~~~}c@{}} 
\includegraphics[width=0.23\textwidth]{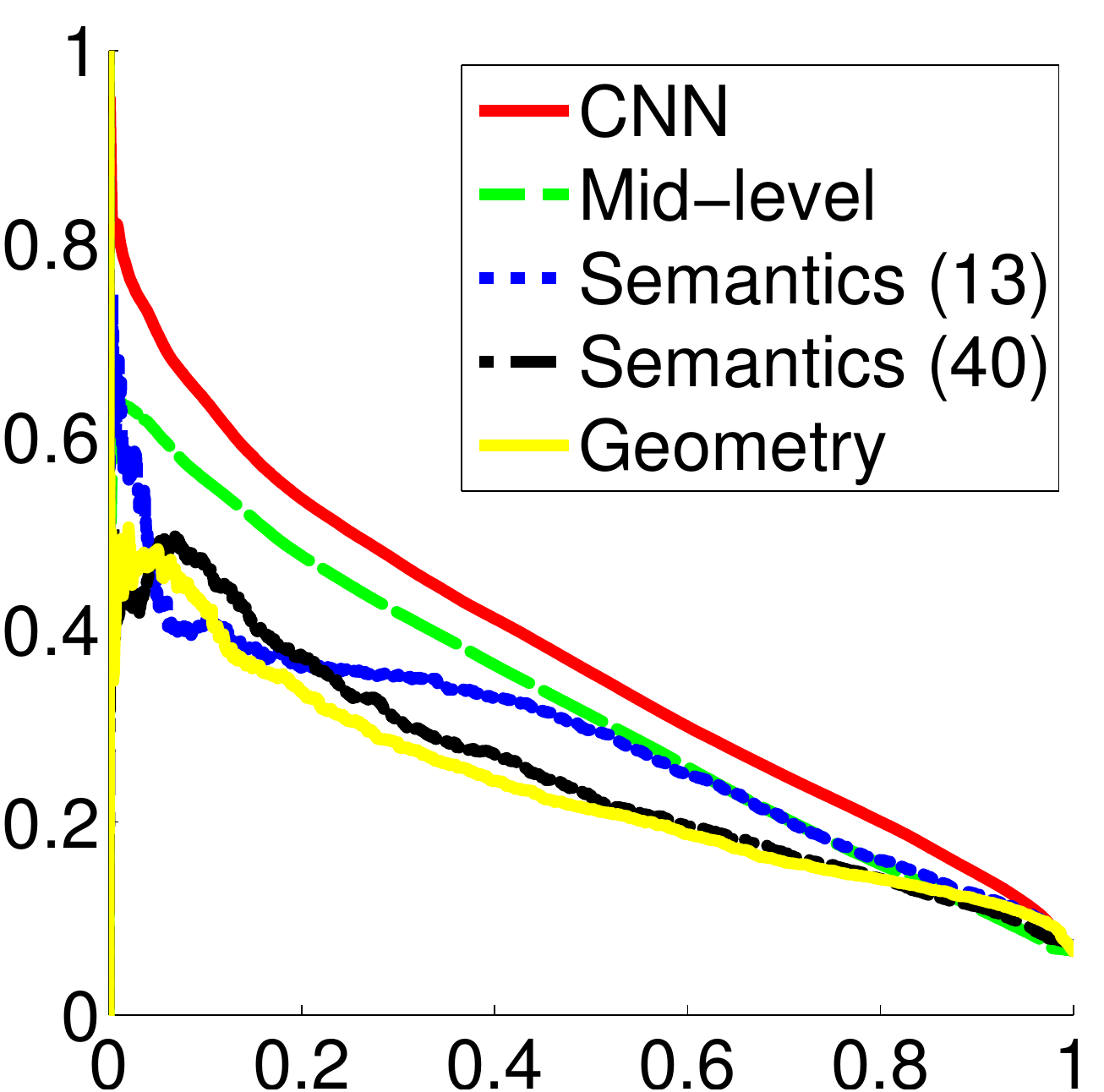} & 
\includegraphics[width=0.23\textwidth]{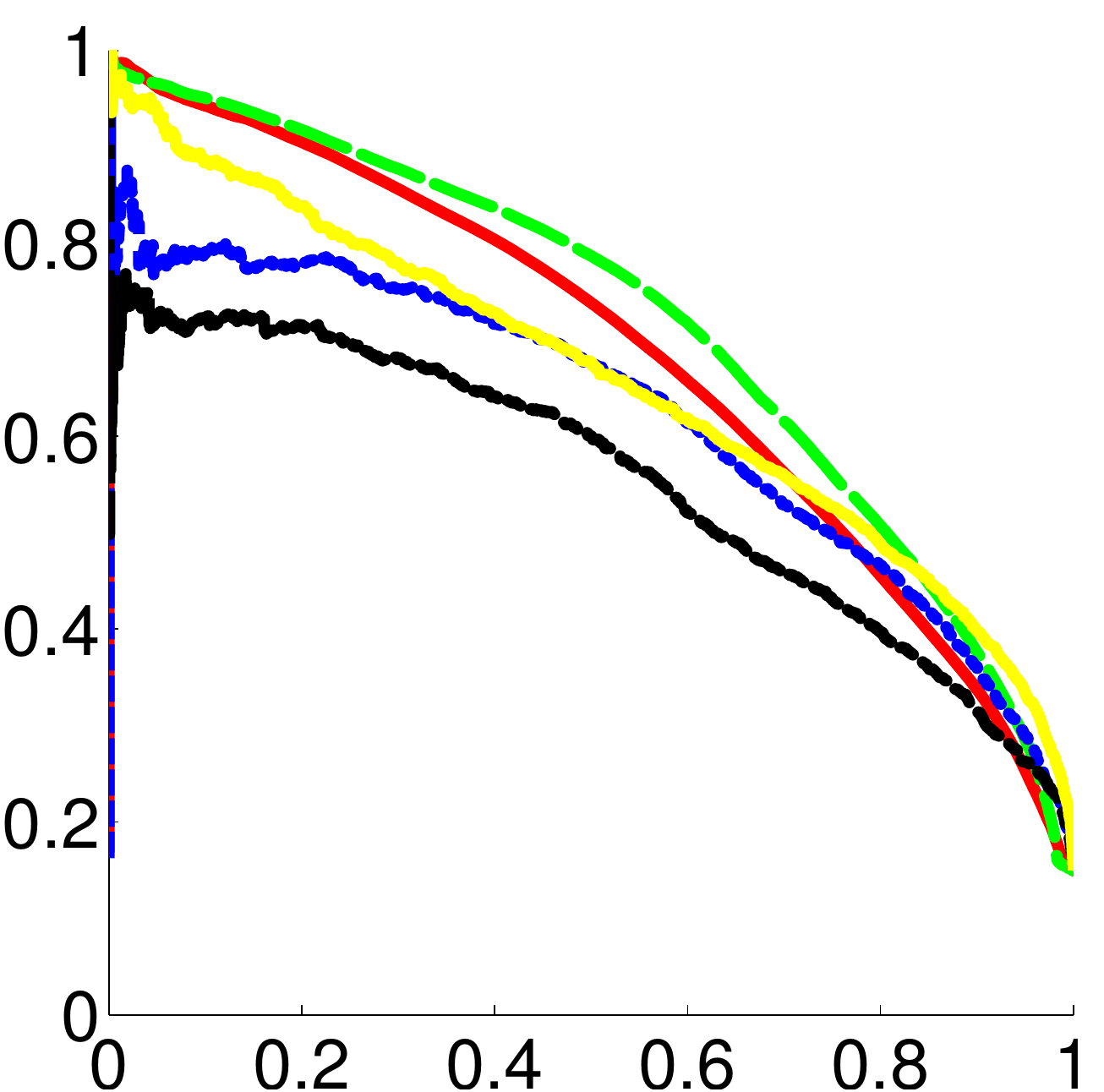} \\
Sitting & Reach (Feet) \\ 
\end{tabular} 
\vspace{-0.1in}
\caption{Precision-Recall Curves for proposed approaches compared to 
mediated approaches.}
\label{fig:pr}
\vspace{-0.0in}
\end{figure}

\subsection{Results on UIUC}

A previous work on affordance estimation from Gupta et al.\ \cite{Gupta11} labeled affordances
in images from the UIUC dataset of \cite{Hedau09}. We evaluate on the
intersection of our label sets -- sitting upright and lying down.
Gupta et al.\ evaluated on only the 25 images of the test set
where the vanishing point estimators used; we expand this to the full set. Even 
though UIUC is very different from NYUv2, we directly apply the models learned on NYUv2. 
To compare with \cite{Gupta11}, we threshold our predictions by choosing the
value that maximizes the Jaccard index on training data.
Our Mid-Level approach achieves the best performance among the direct approaches, getting 
$0.17$ in sitting and $0.46$ in lying. This matches or beats the approach of \cite{Gupta11},
which gets $0.14$ and $0.46$ but which was trained on the same dataset. One advantage of our algorithm that is
ignored by evaluating on masks is that it provides a meaningful score that can be used to trade off
precision and recall, rather than simply a mask.

\subsection{Using Affordances}

Finally, we demonstrate the usefulness of our affordances by using them as a feature for scene understanding. A fully
integrated system is beyond this paper's scope, but we report two proof-of-concept results -- box layout estimation
and horizontal surface detection. Each shows that affordances offers complementary cues to geometry understanding;
this suggests that future work on 3D scene understanding might benefit from treating affordances as a 
full-fledged cue rather than as a byproduct of estimated geometry.

In \cite{Fouhey14c}, affordance reasoning was proposed as a cue for single-view geometry. These affordances were
inferred by watching humans interact with the scene, but our models can predict affordances without seeing
humans. We plug the standing and reach-feet locations predicted by our mid-level affordance technique into the same
machinery. We obtain an improvement of $2.5\%$ over only appearance features, compared to the $3.4\%$ obtained in 
\cite{Fouhey14c} by watching hundreds to thousands of frames of a video.

\begin{figure}
\includegraphics[width=\linewidth]{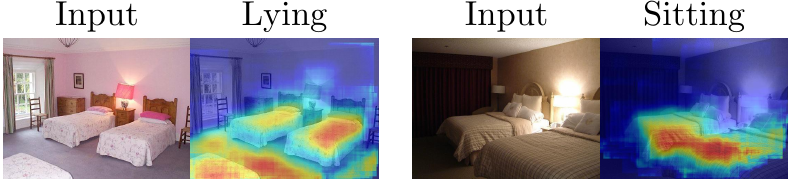}
\caption{Results on the UIUC dataset. We predict lying down on beds and floor (note the bed in the
lower-left corner), and sitting on the edges of beds.}
\vspace{-0.1in}
\end{figure}

We also report results on the task of finding horizontal surfaces in NYUv2 (within $30^\circ$ of the Y-axis). 
If we treat this as a detection task, the estimated $Y$-coordinate from 3DP obtains $70.6\%$ AP;
training a per-pixel random forest on $Y$ and the confidence of 3DP gives $76.9\%$. If we instead just use the 
estimated affordance features at each pixel, we get $80.0\%$, or a gain of $3.1\%$ AP. Combining the confidence and
the affordances yields another $0.3\%$.

%% file: discussion.tex
\section{Conclusion} 

After 35 years of Gibson's theory of affordances, we explore the theory as it was
originally proposed as a tribute to the man who is credited
with one of the biggest ideas in human psychology but who is often discredited when
it comes to the behavioral or algorithmic implications of this theory. 
Specifically, we propose two approaches for the direct perception of affordances;
our experiments show that it may be beneficial to bypass intermediate representations
and that our predictions can aid 3D scene understanding.
We hope this paper rekindles discussion about direct perception.

%% file: direct.bbl
\begin{thebibliography}{10}\itemsep=-1pt

\bibitem{Ba14}
L.~J. Ba and R.~Caruana.
\newblock Do deep nets really need to be deep?
\newblock In {\em NIPS}, 2014.

\bibitem{Castellini11}
C.~Castellini, T.~Tommasi, N.~Noceti, F.~Odone, and B.~Caputo.
\newblock Using object affordances to improve object recognition.
\newblock {\em IEEE Transactions on Autonomous Mental Development}, 3(3), 2011.

\bibitem{Couprie13}
C.~Couprie, C.~Farabet, Y.~LeCun, and L.~Najman.
\newblock Indoor semantic segmentation using depth information.
\newblock In {\em ICLR}, 2013.

\bibitem{Xie13}
S.~T. D.~Xie and S.~Zhu.
\newblock Inferring ‘dark matter’ and ‘dark energy’ from videos.
\newblock In {\em ICCV}, 2013.

\bibitem{Dalal05}
N.~Dalal and B.~Triggs.
\newblock Histograms of oriented gradients for human detection.
\newblock In {\em CVPR}, 2005.

\bibitem{Delaitre12}
V.~Delaitre, D.~Fouhey, I.~Laptev, J.~Sivic, A.~Efros, and A.~Gupta.
\newblock Scene semantics from long-term observation of people.
\newblock In {\em ECCV}, 2012.

\bibitem{Doersch12}
C.~Doersch, S.~Singh, A.~Gupta, J.~Sivic, and A.~A. Efros.
\newblock What makes {P}aris look like {P}aris?
\newblock {\em ACM Transactions on Graphics (SIGGRAPH)}, 31(4), 2012.

\bibitem{Fouhey12}
D.~F. Fouhey, V.~Delaitre, A.~Gupta, A.~A. Efros, I.~Laptev, and J.~Sivic.
\newblock People watching: Human actions as a cue for single-view geometry.
\newblock In {\em ECCV}, 2012.

\bibitem{Fouhey13a}
D.~F. Fouhey, A.~Gupta, and M.~Hebert.
\newblock Data-driven {3D} primitives for single image understanding.
\newblock In {\em ICCV}, 2013.

\bibitem{Fouhey14c}
D.~F. Fouhey, A.~Gupta, and M.~Hebert.
\newblock Unfolding an indoor origami world.
\newblock In {\em ECCV}, 2014.

\bibitem{Gibson79}
J.~Gibson.
\newblock {\em The ecological approach to visual perception}.
\newblock Boston: Houghton Mifflin, 1979.

\bibitem{Grabner11}
H.~Grabner, J.~Gall, and L.~van Gool.
\newblock What makes a chair a chair?
\newblock In {\em CVPR}, 2011.

\bibitem{Gupta07}
A.~Gupta and L.~S. Davis.
\newblock Objects in action: An approach for combining action understanding and
  object perception.
\newblock In {\em CVPR}, 2007.

\bibitem{Gupta11}
A.~Gupta, S.~Satkin, A.~Efros, and M.~Hebert.
\newblock From {3D} scene geometry to human workspace.
\newblock In {\em CVPR}, 2011.

\bibitem{Gupta13}
S.~Gupta, P.~Arbelaez, and J.~Malik.
\newblock Perceptual organization and recognition of indoor scenes from {RGB-D}
  images.
\newblock In {\em CVPR}, 2013.

\bibitem{Hedau09}
V.~Hedau, D.~Hoiem, and D.~Forsyth.
\newblock Recovering the spatial layout of cluttered rooms.
\newblock In {\em ICCV}, 2009.

\bibitem{Jiang13}
H.~Jiang and J.~Xiao.
\newblock A linear approach to matching cuboids in {RGBD} images.
\newblock In {\em CVPR}, 2013.

\bibitem{Kjellstrom08}
H.~Kjellstrom, J.~Romero, D.~Martinez, and D.~Kragic.
\newblock Simultaneous visual recognition of manipulation actions and
  manipulated objects.
\newblock In {\em ECCV}, 2008.

\bibitem{Koppula14}
H.~Koppula and A.~Saxena.
\newblock Physically-grounded spatio-temporal object affordances.
\newblock In {\em ECCV}, 2014.

\bibitem{Koppula15}
H.~Koppula and A.~Saxena.
\newblock Anticipating human activities using object affordances for reactive
  robotic response.
\newblock {\em TPAMI}, 2015.

\bibitem{Krizhevsky12}
A.~Krizhevsky, I.~Sutskever, and G.~E. Hinton.
\newblock Imagenet classification with deep convolutional neural networks.
\newblock In {\em NIPS}, 2012.

\bibitem{Ladicky09}
L.~Ladicky, C.~Russell, P.~Kohli, and P.~H. Torr.
\newblock Associative hierarchical crfs for object class image segmentation.
\newblock In {\em ICCV}, 2009.

\bibitem{Ladicky10}
L.~Ladicky, C.~Russell, P.~Kohli, and P.~H. Torr.
\newblock Graph cut based inference with co-occurrence statistics.
\newblock In {\em ECCV}, 2010.

\bibitem{Ladicky14b}
L.~Ladick\'y, B.~Zeisl, and M.~Pollefeys.
\newblock Discriminatively trained dense surface normal estimation.
\newblock In {\em ECCV}, 2014.

\bibitem{Palmer99}
S.~E. Palmer.
\newblock {\em Vision {S}cience: {P}hotons to {P}henomenology}.
\newblock {MIT} {P}ress, 1999.

\bibitem{Rivlin}
E.~Rivlin, S.~Dickinson, and A.~Rosenfeld.
\newblock Recognition by functional parts.
\newblock In {\em CVIU}, 1995.

\bibitem{Scheirer2012}
W.~J. Scheirer, N.~Kumar, P.~N. Belhumeur, and T.~E. Boult.
\newblock Multi-attribute spaces: Calibration for attribute fusion and
  similarity search.
\newblock In {\em CVPR}, 2012.

\bibitem{Silberman12}
N.~Silberman, D.~Hoiem, P.~Kohli, and R.~Fergus.
\newblock Indoor segmentation and support inference from {RGBD} images.
\newblock In {\em ECCV}, 2012.

\bibitem{Singh12}
S.~Singh, A.~Gupta, and A.~A. Efros.
\newblock Unsupervised discovery of mid-level discriminative patches.
\newblock In {\em ECCV}, 2012.

\bibitem{Stark}
L.~Stark and K.~Bowyer.
\newblock Achieving generalized object recognition through reasoning about
  association of function to structure.
\newblock In {\em PAMI}, 1991.

\bibitem{Ullman80}
S.~Ullman.
\newblock Against direct perception.
\newblock Technical Report AIM-574, MIT A.I. Lab, 1980.

\bibitem{Wang15}
X.~Wang, D.~F. Fouhey, and A.~Gupta.
\newblock Designing deep networks for surface normal estimation.
\newblock 2015.

\bibitem{Binford}
P.~Winston, T.~Binford, B.~Katz, and M.~Lowry.
\newblock Learning physical description from functional definitions, examples
  and precedents.
\newblock In {\em MIT Press}, 1984.

\bibitem{Yao10a}
B.~Yao and L.~Fei-Fei.
\newblock Modeling mutual context of object and human pose in human-object
  interaction activities.
\newblock In {\em CVPR}, 2010.

\bibitem{Zhao13}
Y.~Zhao and S.~Zhu.
\newblock Scene parsing by integrating function, geometry and appearance
  models.
\newblock In {\em CVPR}, 2013.

\bibitem{Zhu14}
Y.~Zhu, A.~Fathi, and L.~Fei-Fei.
\newblock Reasoning about object affordances in a knowledge base
  representation.
\newblock In {\em ECCV}, 2014.

\bibitem{Zhu15}
Y.~Zhu, Y.~Zhao, and S.~Zhu.
\newblock Understanding tools: Task-oriented object modeling, learning and
  recognition.
\newblock In {\em CVPR}, 2015.

\end{thebibliography}
